\documentclass[letterpaper, 10 pt, conference]{ieeeconf}  

\IEEEoverridecommandlockouts                              

\usepackage{amsmath} 

\usepackage{amssymb}  
\usepackage{bbm}
\usepackage[bb=boondox]{mathalfa}
\usepackage{graphicx} 
\usepackage{hyperref}
\usepackage{booktabs}
\usepackage{threeparttable}
\usepackage{tabularx}
\usepackage{multirow}
\usepackage[caption=false, font=footnotesize]{subfig}
\usepackage{algorithmic}
\usepackage{algorithm}

\usepackage{color}
\usepackage{xcolor}
\usepackage{colortbl}
\definecolor{green}{rgb}{0.8,1.0,0.8}
\definecolor{red}{rgb}{1.0,0.8,0.8}
\newcommand{\good}{
  \cellcolor{green}
}
\newcommand{\bad}{
  \cellcolor{red}
}

\usepackage[%
backend=biber,
url=true,
block=space,
maxbibnames=10,
minbibnames=1,
bibstyle=ieee,
]{biblatex}
\AtBeginBibliography{\footnotesize}
\AtEveryBibitem{
  \clearfield{doi}
  \clearfield{isbn}
  \clearfield{issn}
  \clearfield{month}
  \clearfield{publisher}
  \ifentrytype{inproceedings}{
    \clearfield{arxivId}
    \clearfield{archivePrefix}
    \clearfield{eprint}
    \clearfield{url}
  }{}
  \ifentrytype{article}{
    \clearfield{arxivId}
    \clearfield{archivePrefix}
    \clearfield{eprint}
    \clearfield{url}
  }{}
  \ifentrytype{report}{
  }{}
}
\renewbibmacro{in:}{%
  \ifentrytype{inproceedings}{%
    \setunit{}
    \addperiod\addspace In \textit{Proc.\ of the}}%
  {\printtext{\bibstring{in}\intitlepunct}}
}
\DeclareSourcemap{
  \maps{
    \map{ 
      \step[fieldsource=url,
      match=\regexp{\{\\\_\}|\{\_\}|\\\_},
      replace=\regexp{\_}]
    }
    \map{ 
      \step[fieldsource=url,
      match=\regexp{\{\$\\sim\$\}|\{\~\}|\$\\sim\$},
      replace=\regexp{\~}]
    }
    \map{ 
      \step[fieldsource=url,
      match=\regexp{\{\\\x{26}\}},
      replace=\regexp{\x{26}}]
    }
  }
}
\title{\LARGE \bf
Single-Shot Global Localization via \\ Graph-Theoretic Correspondence Matching
}

\author{Shigemichi Matsuzaki$^{1}$, Kenji Koide$^{2}$, Shuji Oishi$^{2}$,
Masashi Yokozuka$^{2}$, and Atsuhiko Banno$^{2}$
\thanks{
*This work was supported in part by a project commissioned by the New Energy and Industrial Technology Development Organization (NEDO).
This work was also supported by JSPS KAKENHI (Grant Number 22K12214).
}
\thanks{$^{1}$S. Matsuzaki is with Department of Computer Science and Engineering, Toyohashi University of Technology,
Hibarigaoka 1-1, Tenpaku-cho, Toyohashi, Aichi, Japan {\tt\small matsuzaki@aisl.cs.tut.ac.jp}}%
\thanks{$^{2}$ K. Koide, S. Oishi, M. Yokozuka, and A. Banno are with
the Department of Information Technology
and Human Factors, the National Institute of Advanced Industrial Science and Technology,
Umezono 1-1-1, Tsukuba, 3050061, Ibaraki, Japan {\tt\small {k.koide}@ieee.org}}%
}

\begin{document}

\maketitle
\thispagestyle{empty}
\pagestyle{empty}


\begin{abstract}
	This paper describes a method of single-shot global localization 
	based on graph-theoretic matching of instances
	between a query and the prior map.
	The proposed framework employs correspondence matching 
	based on the maximum clique problem (MCP).
	The framework is potentially applicable to 
	other map and/or query modalities
	thanks to the graph-based abstraction of the problem,
	while many of existing global localization methods rely on 
	a query and the dataset in the same modality.
	We implement it with a semantically labeled 3D point cloud map,
	and a semantic segmentation image as a query.
	Leveraging the graph-theoretic framework, 
	the proposed method realizes global localization 
	exploiting only the map and the query.
	The method shows promising results 
	on multiple large-scale simulated maps of urban scenes.
\end{abstract}

\section{Introduction}

Global localization is a problem
to estimate a sensor pose in a prior map
given a sensor observation or a sequence of observations,
which is referred to as a \textit{query},
and without prior information of its initial pose.
It is a fundamental ability
of mobile robots and autonomous vehicles
for reliable and safe operation.
For the task of global positioning,
Global Navigation Satellite System (GNSS) is widely used.
It is known, however, that the utility of GNSS
is limited to open space without obstacles
such as buildings, trees etc.
that cause multi-path effect.

Various approaches to global localization
have been proposed in a last few decades.
In traditional Bayesian estimation such as Monte Carlo Localization
(MCL) \cite{Dellaert1999},
a match between sensor readings and the prior map
is evaluated through the observation model
and incorporated as likelihood in
iterative computation of the belief score on the robot pose.
Single-shot methods such as
global descriptor matching \cite{Arandjelovic2018,Khaliq2022}
and local geometry matching \cite{Geppert2019,Dube2020}
instead calculate the most likely pose given a single query.
In the former, a single global descriptor for the query
is computed and used to retrieve from a database
the most similar descriptor associated with a pose.
The latter computes a set of descriptors
that describe the local structure of the query,
and find their correspondence with
pre-computed feature points in the map to compute the pose
via error minimization.

Overall, the global localization methods
essentially require matching between the query and the map.
Regardless of the approaches, many of the existing
global localization methods assume
the query and the map in the same modality,
i.e., the prior map must store data in
the same representation as the query.
This fact limits the applicability of the methods
to a wide variety of different sensor
and map modalities available.
Recently, various map representations have emerged,
such as point cloud maps,
high-definition vector maps (HD maps),
tagged maps like Google Map, etc.
If those cross-modal data can be handled
by a unified method in global localization,
more flexible choices on the sensor and the map
will be available for the global localization problem.

\begin{figure}[tb]
	\centering
	\includegraphics[width=\hsize]{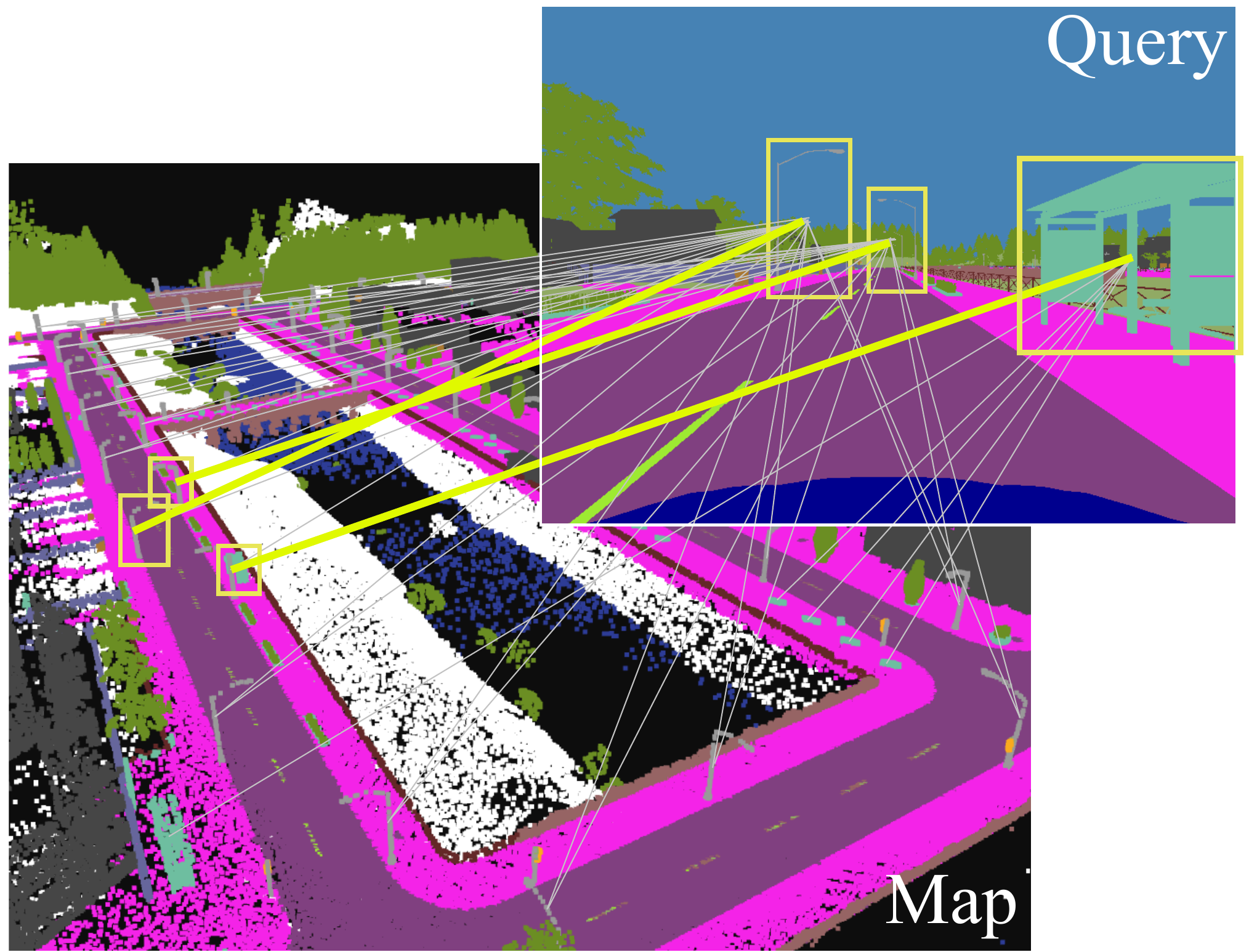}
	\caption{Global localization is solved by estimating
		true correspondences between instances in a query
		and those in the prior map.
		The proposed graph-theoretic framework is
		potentially applicable to
		various query and map modalities.}
	\label{fig:top_image}
\end{figure}

With such a prospect,
we propose a framework of single-shot global localization,
formulating it as
instance correspondence matching between the query and the prior map
and finding a pose in the map that
best explains the correspondences.
For the correspondence matching, we employ
a graph-theoretic method that reduces the problem to
the maximum clique problem (MCP).
Specifically, we construct a graph called
a \textit{consistency graph} that represents local consistency
between correspondence candidates,
and find the most likely set of correspondences via MCP
on the consistency graph, similarly to \cite{Yang2020g} etc.
The motivation behind the formulation is to abstract
the problem to a general graph problem
so that it is not constrained on
a specific modality of the query and the map.
The proposed method can potentially be applied
to cross-modal global localization problems 
as long as a proper consistency criteria can be defined.

In the present work, we implement the framework using
3D semantic point cloud maps as a prior map
and a semantic segmentation image as a query
for correspondence matching.
The consistency of the correspondences are evaluated based on
the closeness in appearance from the pose
calculated by a correspondence pair.
Thanks to the graph-theoretic framework,
global localization is realized
with a combination of simple consistency criteria
that does not require additional data for training.
The method thus exploits only a 3D semantic map
and a semantic segmentation query,
in contrast to recent existing methods \cite{Dube2020,Cramariuc2021}
that depend on data-driven descriptors for correspondence matching.
Our method can, therefore, be easily applied to various environments.


\section{Related Work}

\subsection{Global localization}

There have been various approaches
with different techniques and sensor modalities.

\textbf{Iterative methods}
Traditionally, global localization has been
solved in iterative algorithm such as
Monte Carlo localization (MCL) \cite{Dellaert1999}
where the belief on the robot's pose
gradually converges by incorporating a temporal
sequence of observations as the robot moves.
This approach inherently requires motion of the robot,
which limits the use case of the methods.
In the present work, we opt for a single-shot
global localization approach.

\textbf{Global descriptor matching}
Global descriptor-based methods learn
a discriminative descriptor per scene
from the sensor observations.
The pose is then estimated by
matching the descriptor for the observation
and pre-computed set of descriptors associated
with a pose.
This line of work includes
Visual Place Recognition (VPR)
\cite{Arandjelovic2018,Khaliq2022},
geometry-based methods
\cite{Yin2018,Cop2018},
and hybrid methods \cite{Schonberger2018,Ratz2020d}.

\textbf{Local geometry matching}
In visual global localization,
another popular approach is
matching local visual features
such as SIFT \cite{Lowe2004} and SURF \cite{Bay2006}
between a query image and a pre-computed map
and optimizing the pose based on the feature correspondences
\cite{Geppert2019,Se2005}.
In 3D LiDAR-based geometry matching,
local point descriptors such as \cite{Rusu2009}
and point segments \cite{Dube2020,Cramariuc2021}
are often used.

\textbf{Semantics-based localization}
Following the recent progress of semantic segmentation based on DNNs,
global localization methods based on visual semantic information
have also been actively studied.
Results of semantic segmentation are reported to be
more robust to appearance changes than
low-level visual information such as feature points
\cite{Orhan2022}.
Semantic information has been, therefore, exploited
to realize localization methods that are robust to
appearance changes over time 
and viewpoint changes
\cite{Guo2021a}.

\textbf{DNN-based pose regression}
DNNs are also exploited to directly estimate a sensor pose.
Kendal et al. \cite{Kendall2015} first proposed to
regress a 6 DOF pose directly from an image.
Feng et al. \cite{Feng2019} jointly trained the descriptors
for 2D images and 3D points to match data
in those different modalities.
This approach, again, is specialized to the data
on which the network is trained
and hard to apply to various scenes.

\textbf{Problems in existing work and novelties of our work}
Regardless of the approaches,
most of the conventional methods
use a query and a map in the same modality.
On the other hand, the proposed method
does not strictly assume the same modality.
Our method, therefore, is potentially
applicable to various types of maps
such as point cloud maps and HD vector maps.

Although some recent works use DNNs
to learn a strong global/local descriptor
\cite{Arandjelovic2018,Dube2020},
or to embed data from different modalities
in a common feature space \cite{Feng2019},
such methods require a huge amount of training data.
Moreover, pose regression using DNN
is specialized to the data on which the model is trained,
and difficult to apply to various scenes.

Our primary purpose is
to develop a novel unified framework for global localization
applicable to cross-modal settings.
The pose is efficiently estimated
using the graph-theoretic correspondence matching
and simple criteria of appearance similarity.
We show the effectiveness in an implementation
using a semantic point cloud map and a semantic segmentation image.

\subsection{Correspondence matching}

Correspondence matching is a crucial task for
some of the aforementioned methods.
%
Random sample consensus (RANSAC) is
traditionally used to extract
correspondences out of candidates including outliers
\cite{Se2005,Geppert2019}.
Although it shows a high robustness against
false positive matches, 
the number of iterations to achieve
a reasonable result increases exponentially
as a noise ratio increases
due to its stochasticity,
and thus is not applicable
to cases of high outlier rates.

Graph-theoretic correspondence extraction
has been recently shown effective
under a high outlier rate where
RANSAC fails to extract correspondences
\cite{Yang2020g}.
Bailey et al. \cite{Bailey2000a} was
one of the earliest to frame data association
as MCP.
The point association via MCP has been applied to
global registration of point cloud \cite{Yang2020g}
\cite{Shi2021}.
Koide et al. \cite{Koide2022}
proposed an automatic localization of fiducial
tags on a 3D prior map using MCP-based
data association.

To the best of our knowledge,
this work is the first attempt
to employ MCP-based correspondence matching
for global localization.
The proposed method leverages the graph-theoretic approach
to efficiently find the most likely instance correspondences
and carries out stable global localization.

\section{Global Localization via Correspondence Matching by MCP}


%

The general algorithm of the proposed framework
of global localization
is shown in Algorithm \ref{alg1}.
The core of the framework is to estimate the correspondences
between the query instances and the map instances
to calculate the pose that matches the correspondences the best.
To this end, we employ graph-theoretic correspondence matching.

First, we generate candidates for correspondences
between a query instance and a map instance.
Here, we treat all possible pairs of an instance in the query
and one in the map as valid candidates for
correspondences.

Once the correspondence candidates are generated,
we build a \textit{consistency graph}, 
a node of which represents
a candidate of correspondence between
an instance in the query and one in the map,
and an edge indicates consistency
between two candidates represented
by the nodes connected by the edge.
All possible pairs of two correspondence candidates
are evaluated based on certain consistency criteria,
and if two candidates are evaluated as consistent,
an edge is added between the nodes.

The correspondences are then estimated by solving MCP on
the consistency graph,
and the robot pose is calculated using
the extracted correspondences
so that it best satisfies the observations of the instances.

%

%


\begin{algorithm}[tb]
	\caption{Global localization via graph-theoretic correspondence matching}
	\label{alg1}
	\begin{algorithmic}[1]
		\REQUIRE Map instance set $\mathcal{M}=\left\{m_i\right\}^{N_m}_{i=1}$, query instance set $\mathcal{Q}=\left\{q_j\right\}^{N_q}_{j=1}$
		\ENSURE 3-DOF robot pose $\left(x, y, \theta\right)$
		\STATE Correspondence candidate set $\mathcal{C}=\mathcal{M}\times \mathcal{Q}$
		\STATE Vertex set $\mathcal{V}=\mathcal{C}$
		\STATE Edge set $\mathcal{E}=\mathbb{0}$
		\FOR{every combination of $c_i, c_j\in\mathcal{C}$}
		\IF{is\_consistent$\left(c_i, c_j\right)$}
		\STATE $\mathcal{E}\leftarrow \mathcal{E}\cup \left\{e_{ij}, e_{ji}\right\}$
		\ENDIF
		\ENDFOR
		\STATE $\mathcal{C}^{\ast}\leftarrow$max\_clique$\left(\mathcal{V}, \mathcal{E}\right)$
		\STATE $\left(x, y, \theta\right)\leftarrow$calculate\_pose$\left(\mathcal{C}^{\ast}\right)$
		\RETURN $\left(x, y, \theta\right)$
	\end{algorithmic}
\end{algorithm}

\section{Implementation on 3D Semantic Map and a Semantic Segmentation Image}
\label{sec:general_algorithm}

\begin{figure*}[tb]
	\centering
	\includegraphics[width=0.90\hsize]{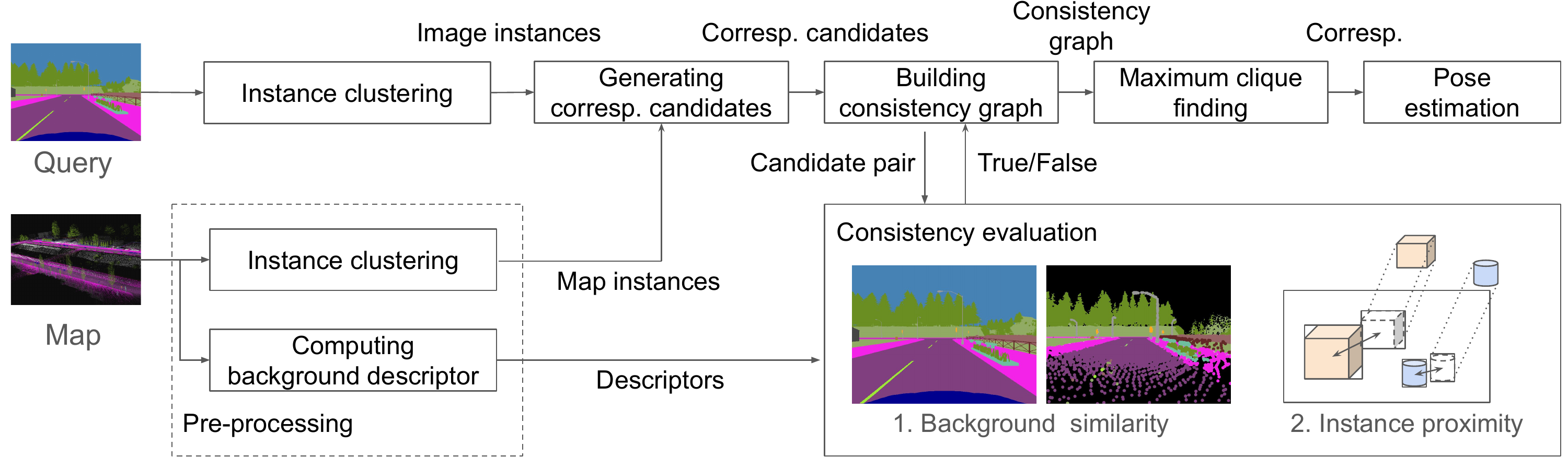}
	\caption{Overview of the proposed framework.
		As pre-processing, the map instances are clustered,
		and the grid histogram descriptors
		(see \ref{sec:impl_consistency_bg_sim})
		are computed, and stored as a database.
		Given a query image, instances are clustered
		and correspondence candidates are generated
		as all possible combinations of a query and a map instances.
		For every pair of the candidates,
		a pose is calculated (\ref{sec:impl_consistency_pose_calculation})
		and the consistency is evaluated
		based on two criteria 1) background similarity, and
		2) instance proximity to construct a consistency graph
		(\ref{sec:impl_consistency_criteria}).
		The most likely set of correspondences is yielded by
		solving MCP on the consistency graph,
		and the pose is calculated based on it
		(\ref{sec:impl_final}).
	}
	\label{fig:method_overview}

\end{figure*}

Here, we implement the proposed framework 
using 3D labeled point cloud map as a prior map,
and a semantic segmentation image as a query.
We consider estimation of a 3 DOF pose
$(x, y, \theta)$ of the ground robot.
We assume a robot with a camera facing forward
and that its transformation relative to the robot's base is known,
and a semantic segmentation image for an image taken by
the camera is available.
The overview of the workflow is shown in Fig. \ref{fig:method_overview}.

\subsection{Clustering the map and the query instances}

The instances of the 3D map are clustered and individually saved
before the localization process.
Here, the following four object classes are used for correspondence matching:
\textit{pole, traffic sign, traffic light, static}.

The map points are simply clustered based on their object class and
spacial closeness.
For the query image, the instances are clustered
based on the connectivity.
This may wrongly extract multiple instances overlapping in the image as one.
We, however, treat such ``connected'' instances as an individual instance,
expecting that they will be rejected in the subsequent process
while there still remain sufficient correspondences for pose estimation.
Candidates of instance correspondences are
all patterns of pairs of instances with the same object class
between the map and the image.

\subsection{Consistency evaluation}

For every combination of two correspondence candidates,
a pose is calculated
so that it matches the size and
the location of image instances
and reprojection of the corresponding map instances.
The consistency of the pair
is then evaluated based on
the similarity of the appearance
between the query and the map reprojection from
the calculated pose.

\subsubsection{Pose calculation using a correspondence pair}
\label{sec:impl_consistency_pose_calculation}

\begin{figure}[tb]
	\centering
	\subfloat[Angle defined by image instances \label{fig:angle_in_image}]{
		\includegraphics[width=0.48\hsize]{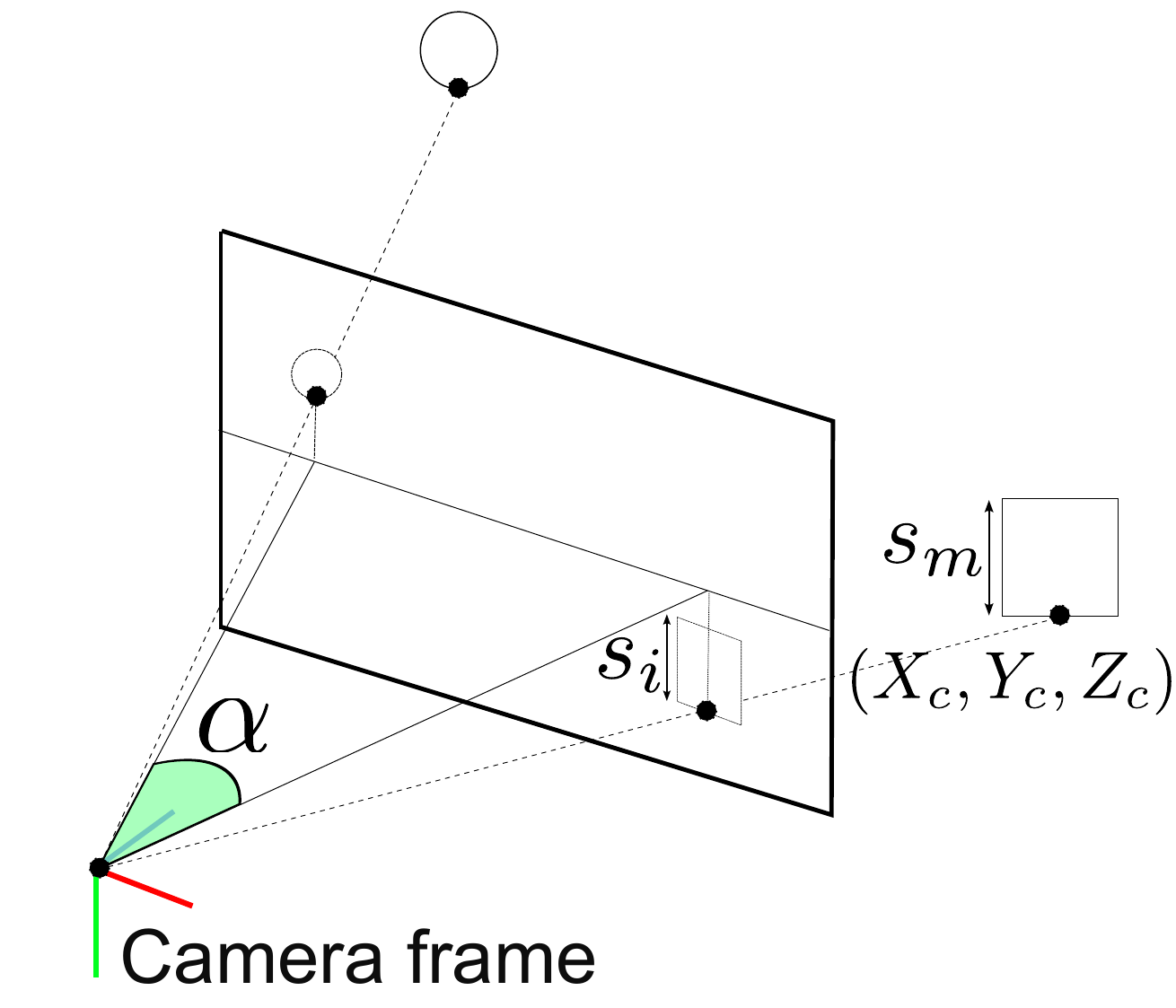}}
	\subfloat[Circle $C$ defined by inscribed angle $\alpha$ \label{fig:how_to_calculate_by_circle}]{
		\includegraphics[width=0.32\hsize]{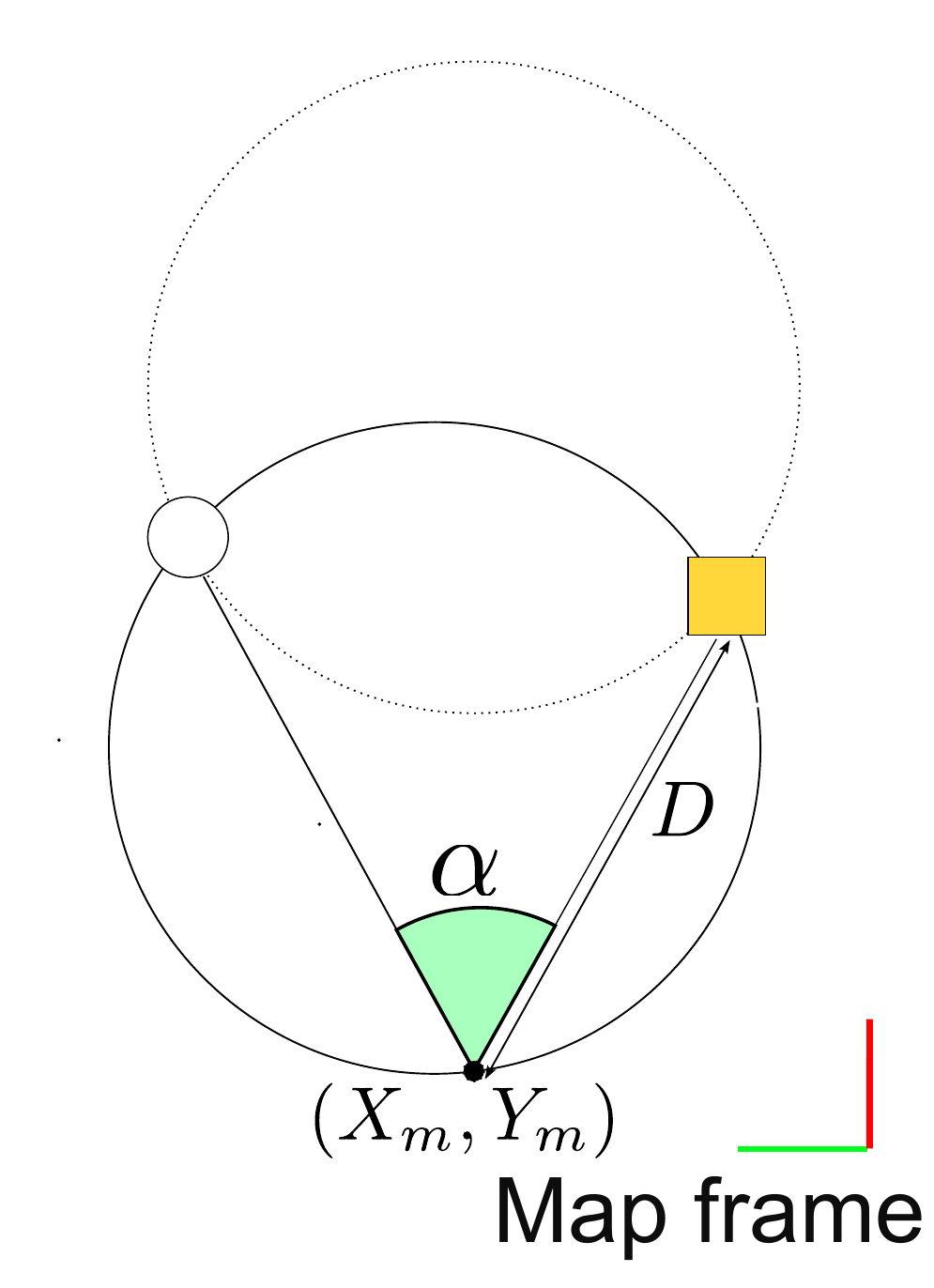}}
	\caption{The candidate regions of the robot pose based on
		the inscribed angle and the size of the instance.
		(b) The translation of the camera $(X_m, Y_m)$
		in the map frame is determined as a point on
		$C$ where the distance to an object is $D=\sqrt{X_c^2 + Z_c^2}$.}
	\label{fig:pose_calculation}
\end{figure}


Given two image instances, an angle $\alpha$
formed by two vectors to each image instance
from the origin of the camera frame is determined.
The region of the pose
is then constrained to a circle $C$ that has
an inscribed angle $\alpha$
as shown in Fig. \ref{fig:pose_calculation}.
Since there are two such circles,
we determine $C$ 
based on the positional relationship of the instances.

To determine the pose,
we use the scale information of one of the correspondences.
The relationship of the size in the 3D space $s_m$
and that of its projection $s_i$ is written as:
\begin{eqnarray}
	s_i = \frac{f\cdot s_m}{Z_c},
	\label{eq:scale_relation}
\end{eqnarray}
where $f$ denotes the focal length of the camera
and $Z_c$ denotes the $Z$ coordinate of the instance
in the camera frame.
Since the sizes of both image instance and the map instance
in a correspondence are known,
$Z_c$ can be estimated as:
\begin{eqnarray}
	Z_c = \frac{f\cdot s_m}{s_i}.
	\label{eq:distance_from_instance}
\end{eqnarray}
The set of points that satisfy
eq. \eqref{eq:distance_from_instance}
also forms a circle with a radius $D=\sqrt{X_c^2 + Z_c^2}$.
The camera pose can, therefore,
be calculated as an intersection
of circle $C$ and a circle that centers at the location of
the map instance and has radius $D$.
In this way, at most two intersections are yielded.
In such a case, we calculate the expected
size of the projection of another map instance 
using eq. \eqref{eq:scale_relation}
from the two poses, and compare them with the size of
the corresponding image instance.
The pose that provides a smaller error of the size
is selected as the solution.

\subsubsection{Consistency criteria}
\label{sec:impl_consistency_criteria}

The consistency is evaluated
based on two criteria, \textit{background similarity}
and \textit{instance proximity}.
The latter is relevant to
the spacial consistency, which is often employed
in MCP-based correspondence matching
such as \cite{Yang2020g} and \cite{Koide2022}.
In our problem, however,
this criterion is not sufficient
because there may be many wrong correspondences
that happen to satisfy it.
The background similarity criterion
is, therefore, employed before the instance proximity
to filter out unlikely correspondences
based on the appearance similarity.
Using the two criteria,
we effectively evaluate
consistency of the correspondences.

\begin{figure}[tb]
	\centering
	\subfloat[Grid histogram \label{fig:consistency_criteria_grid_histogram}]{
		\includegraphics[width=0.50\hsize]{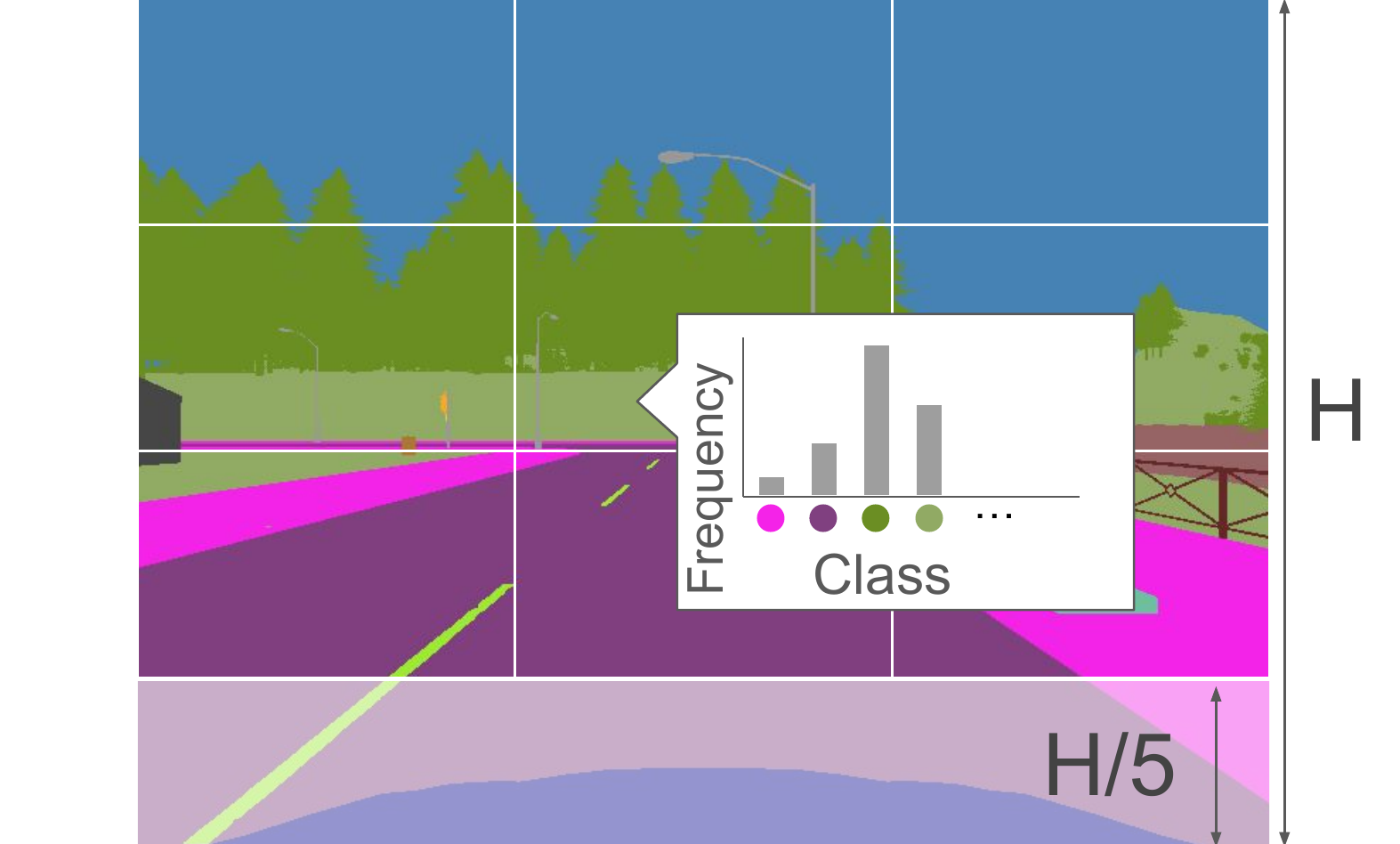}}
	\subfloat[Instance proximity \label{fig:consistency_criteria_instance_proximity}]{
		\includegraphics[width=0.50\hsize]{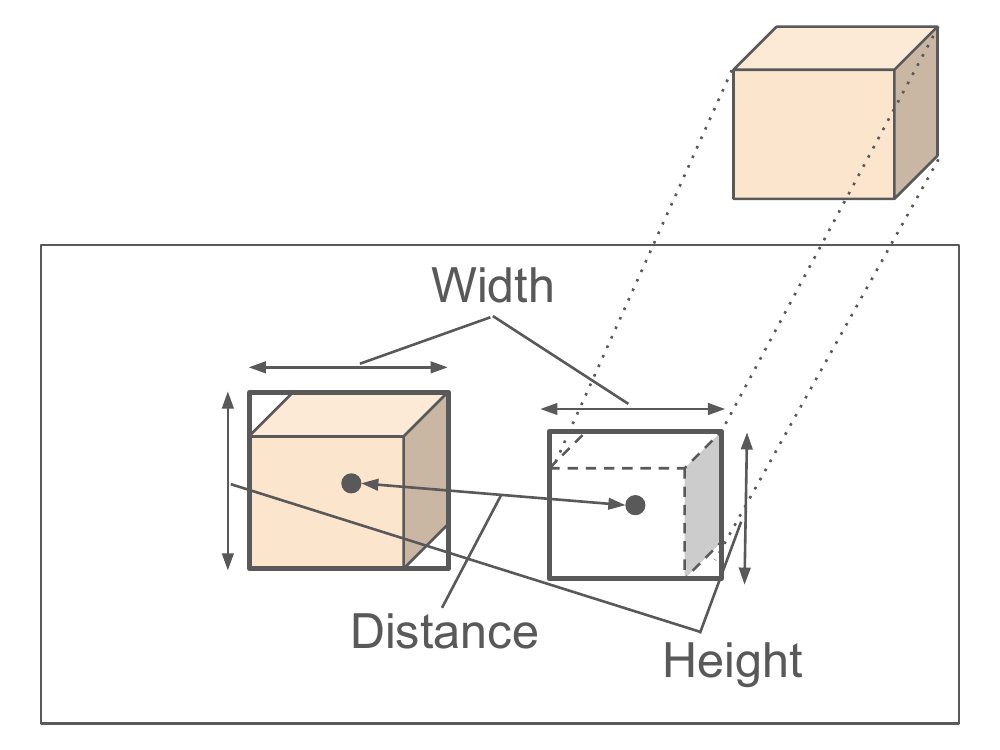}}
	\caption{Consistency criteria.
		(a) The grid histogram descriptor is a normalized vector
		of the local histograms of object classes.
		(b) The instance proximity
		evaluates a spacial relationship between
		the bounding boxes around the image instance
		and the projection of the map instance.}
	\label{fig:consistency_criteria}
\end{figure}

\textbf{Background similarity}
\label{sec:impl_consistency_bg_sim}
The purpose of the background similarity criterion
is not to strictly
detect images similar to the query,
but to quickly filter out obvious outliers.
To this end, we employ a simple histogram descriptor.

Fig. \ref{fig:consistency_criteria_grid_histogram}
shows an overview. 
First, the bottom of the image is cut off as
this part 
does not convey much information.
The remaining part of the image is
divided into grids with a fixed size.
For each grid, a histogram of the object classes is generated
and L2-normalized.
We chose nine background object classes:
\textit{building, fence, road line,  road,  sidewalk, vegetation, wall, bridge, terrain}.
The normalized histograms from the grids are concatenated
into ($h\times w \times 9$)-dimensional vector,
where $h$ and $w$ denotes the number of grid
in the vertical and the horizontal axis, respectively.
The vector is then L2-normalized again
to form a descriptor. 
Similarity of two descriptors are given as a dot product,
or equivalently a cosine similarity with
a value range of $[0, 1]$.

For fast evaluation, we pre-compute the descriptors for the map
by projecting the 3D map from the poses
uniformly sampled in the map.


\textbf{Instance proximity}
\label{sec:impl_consistency_inst_sim}
The instance proximity is evaluated by the distance and the size differences
of bounding boxes around the projected map instance
and the image instance
(see Fig. \ref{fig:consistency_criteria_instance_proximity}).
The distance between them are defined as
the Euclidean distance between the centroids
of the bounding boxes.
The size differences are calculated with 
height and width lengths of the bounding boxes. 
If the distance, the height and the
width difference are all less than a threshold
for both of the correspondences in the given pair,
the pair of correspondence candidates is
considered consistent.

Here, we set the threshold adaptively to the $Z$
coordinate value $Z_c$ of each map instance in the camera frame.
We empirically define the distance threshold $\theta_d$
and the size threshold $\theta_s$
as follows:


\begin{equation}
	\begin{aligned}
		\theta_d & =\min\left(m_d, \max\left(m_{min}, dm_d/Z_c\right)\right)  \\
		\theta_s & =\min\left(m_s, \max\left(m_{min}, dm_s/Z_c\right)\right),
	\end{aligned}
\end{equation}
where $m_d$ and $m_s$ denote the maximum possible threshold
of the instance distance and the size in pixel, respectively.
The idea behind these definitions is to set the thresholds of
objects at $d$ [m] from the sensor to $m_d$ [px] and $m_s$ [px],
and decrease it as $Z_c$ becomes larger.
The thresholds are clipped
with the minimum value of $m_{min}$ [px] and the maximum value of $m_d$ and $m_s$.
Here, we set $d$ to 8 and $m_{min}$ to 5.


\subsection{Pose estimation}
\label{sec:impl_final}

A consistency graph is constructed based on
the aforementioned consistency criteria,
and the most likely set of correspondences are
yielded by solving
as described in \ref{sec:general_algorithm}.
The final pose is then calculated by minimizing
the reprojection error of representation points
of the correspondences.
A representation point of an image instance
is a point that has the largest y value
in the image coordinate,
and that of a map instance is the
point with the smallest z value in
the map coordinate,
representing their bottom point.
Although this way 
does not provide exact matching of
the parts of the instances,
it does not affect much the calculation of the camera pose.
An initial pose is given as an average of
poses calculated by the correspondences in the clique
during the phase of building the consistency graph
(see \ref{sec:impl_consistency_pose_calculation}).
The Levenberg-Marquardt algorithm \cite{Levenberg1944,Marquardt1963}
with Huber kernel \cite{Huber1964}
is applied to the pose optimization.

\subsection{Pose verification via correspondence check}
\label{sec:pose_verification}

Ideally, the maximum clique of the consistency graph
represents a set of correct correspondences.
In practice, however, we found that it is often not the case
due to rejection of the correct correspondences
mainly caused by error in pose calculation,
while evaluating wrong ones as consistent.

To mitigate the problem, we introduce a verification step.
After constructing the consistency graph,
we calculate multiple correspondence set
by iteratively applying MCP solver,
calculating a pose with each set,
and removing edges between the vertices corresponding to the members
of the found maximum clique.
To verify each correspondence set, 
we count the number of image instances
that have a map instance projected close to it
from the calculated pose
based on the proximity criterion.
The correspondence set with the maximum count is treated as the result.
Top N results are yielded by taking results with five largest counts.
\section{Experiments}
\label{sec:experiments}

\subsection{Experimental setup}

The experiments were conducted on
a laptop with an Intel Core i7-10750H (12 threads)
and 16 GB of RAM.


We used simulated point cloud maps of Town01 to 05
from CARLA simulator \cite{Dosovitskiy2017} for evaluation.
The description of the environments is shown in Table \ref{table:carla_towns}.
We created point cloud maps using
the scripts shared by the authors of \cite{Deschaud2021}.
For each map, descriptors are pre-computed from
the poses uniformly sampled
with a stride of 2 m in the x and the y axes of the map frame
and $30^\circ$  in the yaw angle.

The samples of query were generated
from the spawn points of vehicles defined in CARLA.
A camera was mounted on the simulated vehicle
at the height of approx. 1.5 m
from the robot's base.
Since the proposed method estimates a 3 DOF pose
and assumes fixed height of the camera,
we choose the poses with
the height between 1.46 m and 1.90 m as the samples.
We use ground truth semantic segmentation images
generated by CARLA.

\begin{table}[tb]
	\centering
	\caption[Caption for LOF]{Descriptions of the CARLA Towns\protect\footnotemark}
	\label{table:carla_towns}
	\begin{tabular}{ c l }
		\toprule
		Town & Summary                   \\
		\midrule
		01   & \begin{tabular}{l}A basic town layout consisting of "T junctions".\end{tabular} \\
		02   & \begin{tabular}{l}Similar to Town01, but smaller. \end{tabular} \\
		03   & \begin{tabular}{l}
			The most complex town, with a 5-lane junction, \\
			a roundabout, unevenness, a tunnel, and more.\end{tabular} \\
		04   & \begin{tabular}{l}An infinite loop with a highway and a small town. \end{tabular} \\
		05   & \begin{tabular}{l}
			Squared-grid town with cross junctions and a bridge. \\
			It has multiple lanes per direction.\end{tabular} \\
		\bottomrule
	\end{tabular}
\end{table}
\footnotetext{The descriptions are from \url{https://carla.readthedocs.io/en/latest/core_map/}}

The thresholds of the background similarity (BS),
the instance distance and the instance size in
the instance proximity (IP)
are empirically set to 0.9, 110, 50, respectively.

\subsection{Baselines}
As baselines, we use brute-force descriptor search (BF)
and RANSAC-based correspondence matching.
The former searches for the pre-computed descriptor
that has the highest similarity score with the query
and returns the discretized pose that the descriptor is associated with.
In the latter, RANSAC is used for correspondence finding
instead of the proposed graph-theoretic approach.
For a randomly sampled pair of correspondences,
a pose is calculated as described in \ref{sec:impl_consistency_pose_calculation}
and the number of the correspondences consistent with the pose is counted.
We use the same background similarity threshold as
the proposed method to reject the pair,
and the instance proximity threshold to find the correspondences.
The number of iteration is set to 50000.

\subsection{Evaluation conditions}
We evaluate the localization performance
using three conditions: \textit{$\pm$5}, \textit{$\pm$10},
and \textit{front drift}.
For all of the conditions, the threshold of
yaw error is $\pm 30^\circ$.
\textit{$\pm$5} denotes the case where
the lateral and longitudinal errors are within $\pm 5$ m.
Likewise, \textit{$\pm$ 10} means within $\pm$ 10 m of
lateral and longitudinal errors.
\textit{Front drift} denotes the case where
the error in the x axis of the robot's frame is within $\pm 200$ m
and $\pm 5$ m in the y axis. 
This is typically the case when the estimated pose is on
the same road as the ground truth and facing the same direction,
but is drifted in the longitudinal direction in the robot frame.

\subsection{Localization results}

\begin{table}[tb]
	\centering
	\caption{Results of global localization}
	\label{table:experiment_comparative}
	\begin{threeparttable}
		\begin{tabularx}{\hsize}{@{} *7{>{\centering\arraybackslash}X}@{}}
			\toprule
			Town                              &                     & 01             & 02             & 03       & 04            & 05       \\
			\midrule
			Samples                           &                     & 254            & 101            & 197      & 259           & 234      \\
			\midrule
			\multirow{6}{*}{\textit{$\pm$5}}  & \multirow{2}{*}{T1} & 133            & 60             & 63       & \bad 36       & 76       \\
			                                  &                     & (52.4\%)       & (59.4\%)       & (32.0\%) & \bad (13.7\%) & (32.5\%) \\
			                                  & \multirow{2}{*}{T3} & 157            & 68             & 73       & \bad 45       & 93       \\
			                                  &                     & (61.8\%)       & (67.3\%)       & (37.1\%) & \bad (17.4\%) & (39.7\%) \\
			                                  & \multirow{2}{*}{T5} & 167            & \good 71       & 79       & \bad 48       & 98       \\
			                                  &                     & (65.7\%)       & \good (70.3\%) & (40.1\%) & \bad (18.5\%) & (41.9\%) \\
			\midrule
			\multirow{6}{*}{\textit{$\pm$10}} & \multirow{2}{*}{T1} & 142            & 65             & 72       & \bad 49       & 81       \\
			                                  &                     & (55.9\%)       & (64.4\%)       & (36.5\%) & \bad (18.9\%) & (34.6\%) \\
			                                  & \multirow{2}{*}{T3} & 170            & \good 74       & 85       & \bad 62       & 100      \\
			                                  &                     & (66.9\%)       & \good (73.3\%) & (43.1\%) & \bad (23.9\%) & (42.7\%) \\
			                                  & \multirow{2}{*}{T5} & \good 183      & \good 77       & 94       & \bad 67       & 106      \\
			                                  &                     & \good (72.0\%) & \good (76.2\%) & (47.7\%) & \bad (25.9\%) & (45.3\%) \\
			\midrule
			\multirow{6}{*}{\textit{FD}}      & \multirow{2}{*}{T1} & 175            & \good 75       & 95       & \bad 53       & 87       \\
			                                  &                     & (68.9\%)       & \good (74.3\%) & (48.2\%) & \bad (20.5\%) & (37.2\%) \\
			                                  & \multirow{2}{*}{T3} & \good 196      & \good 81       & 104      & \bad 76       & 110      \\
			                                  &                     & \good (77.2\%) & \good (80.2\%) & (52.8\%) & \bad (27.8\%) & (47.0\%) \\
			                                  & \multirow{2}{*}{T5} & \good 207      & \good 86       & 114      & 82            & 123      \\
			                                  &                     & \good (81.5\%) & \good (85.1\%) & (57.9\%) & (30.1\%)      & (52.6\%) \\
			\bottomrule
		\end{tabularx}
	\end{threeparttable}
	\begin{tablenotes}[flushleft]
		\item * T1, T3, and T5 denote Top 1, Top 3, and Top 5, respectively. \textit{FD} denotes \textit{Front drift}. \colorbox{green}{Green}: $> 70$ \%, \colorbox{red}{Red}: $< 30$ \%
	\end{tablenotes}
\end{table}

Table \ref{table:experiment_comparative} shows the results
of global localization.
The values indicate the number of samples
that resulted in fulfilling the evaluation conditions
in Top 1, Top 3, and Top 5 results calculated
as described in \ref{sec:pose_verification}.
In Town01 and 02, Top 5 estimation for
approximately 65 to 70 \% of the samples
fulfilled \textit{$\pm$5},
and more than 80 \% of them fulfilled \textit{front drift}.
In Town 03 and 05, Top 5 results of
approximately 40 \% samples fulfilled \textit{$\pm$5}.
The result of Town04 was significantly worse than the other maps,
with 18.5 \% of \textit{$\pm$5}
and 30.1 \% of \textit{front drift} in Top 5 results, respectively.
We discuss the cause of the low performance in \ref{sec:discussion}.

\subsection{Comparison with the baselines}

Table \ref{table:experiment_baseline} shows the comparative results
with the baselines.
The values are the number of Top 1 results
of \textit{$\pm$5} and \textit{$\pm$10}.
The proposed method surpassed the baseline methods in
all conditions but \textit{$\pm$10} of BF in Town05.
It is worth noting that the brute-force search
did not result in accurate pose estimation
in most of the environments.
Nevertheless, combined with our framework, it resulted in
better performance. 
RANSAC was also resulted in worse results than the proposed method.
This is an expected result as
most of the correspondence candidates are outliers in the tasks,
and RANSAC struggles with finding a consistent pair
via random sampling.
In contrast, the proposed method efficiently found
the correspondences leveraging the graph-theoretic approach.

%
\begin{table}[tb]
	\centering
	\caption{Comparison with the baselines}
	\label{table:experiment_baseline}
	\begin{tabular}{ c c c c c c c }
		\toprule
		                                   & Town no.         & 01           & 02          & 03          & 04          & 05          \\
		\cline{2-7}
		                                   & Samples          & 254          & 101         & 197         & 259         & 234         \\
		\midrule
		\multirow{2}{*}{BF}                & \textit{$\pm$5}  & 99           & 35          & 49          & 32          & 70          \\
		                                   & \textit{$\pm$10} & 107          & 34          & 62          & 47          & \textbf{98} \\
		\midrule
		\multirow{2}{*}{RANSAC}            & \textit{$\pm$5}  & 17           & 7           & 5           & 2           & 6           \\
		                                   & \textit{$\pm$10} & 22           & 10          & 17          & 11          & 12          \\
		\midrule
		\multirow{2}{*}{\textbf{Proposed}} & \textit{$\pm$5}  & \textbf{133} & \textbf{60} & \textbf{63} & \textbf{36} & \textbf{76} \\
		                                   & \textit{$\pm$10} & \textbf{142} & \textbf{65} & \textbf{72} & \textbf{49} & 81          \\

		\bottomrule
	\end{tabular}
\end{table}

\subsection{Ablation studies on the consistency criteria}

\begin{table}[tb]
	\centering

	\caption{Ablation on the consistency criteria}
	\label{table:experiment_ablation_consistency}
	\begin{threeparttable}
		\begin{tabular}{ c c c c c c c c }
			\toprule
			Town no.      & 01           & 02          & 03          & 04          & 05          \\
			\midrule
			Samples       & 254          & 101         & 197         & 259         & 234         \\
			\midrule
			BS(0.9)       & 40           & 18          & 6           & 1           & 8           \\
			IP/IP+BS(0.5) & 18*          & 8           & 4           & 11          & 15          \\
			IP+BS(0.9)    & \textbf{133} & \textbf{60} & \textbf{63} & \textbf{36} & \textbf{76} \\
			\bottomrule
		\end{tabular}

		\begin{tablenotes}[normal] 
			\item[*] Only IP
		\end{tablenotes}
	\end{threeparttable}
\end{table}

Table \ref{table:experiment_ablation_consistency} shows
the results of an ablation study on the consistency criteria
to examine their contribution. 
To test IP for Town02 to 05,
we loosened BS threshold to 0.5
(denoted as IP+BS(0.5))
instead of using only IP,
because we found disabling BS
significantly slowed down the computation speed
due to increased number of transformation of
map instances in IP computation.  

As a result, the performance was significantly worse
in BS and IP (Town01) / IP+BS(0.5) (02-05).
The combination of those simple consistency criteria
resulted the best to effectively find
the correspondences and calculate the pose.


\subsection{Discussion}
\label{sec:discussion}

We saw promising results of the proposed framework
especially in Town01 and 02.
Notably, the proposed method requires only
the semantic 3D map and semantic segmentation query.
Although the building blocks for
consistency evaluation are simple,
our system provided reasonable results.
We suppose this is thanks to
the graph-theoretic correspondence matching approach
which is capable of effectively extracting correspondences
under high outlier rate.


\textbf{Limitations}
Despite the powerful graph-theoretic approach,
there were also many failure cases.
They 
mainly stem from
the error in the pose estimation using a pair of
correspondences.
The current pose calculation is sensitive to noise
in instance clustering because
it exploits the relationship of the scales of
instances to estimate the distance from
the robot pose to the map instances,
and assumes that
both the image instance and the map instance are complete.
In reality, however, the assumption often does not hold.
This problem will also affect real-world application
because, in practice, the query semantic segmentation will be
provided by an estimator based on DNNs etc.,
which inevitably has estimation noise.
To deal with it,
we should introduce pose calculation
that is more robust to such noise.

The causes of the low accuracy for Town04 are twofold.
First, the number of image instances are 
fewer 
than other Towns.
Due to the unstable pose calculation,
a few image instances in an image are generally rejected in MCP,
and it is thus more possible to result in
no correct correspondences detected when
an image has few instances.
Second,
the mountains and terrains seen far away in the query
were not mapped
due to the limited range of the point cloud map,
and the descriptors of the query
and the map had a large difference
in many samples.
This effect was especially significant in Town04
consisting of open highway scenes with mountains.
Although the simple descriptor provided
reasonable results in tandem with the proposed framework,
we should reconsider its design
to mitigate such a problem.


\section{Conclusion}

We proposed a framework for global localization
based on graph-theoretic matching of
instance correspondences
between a query and a prior map.
We implemented the framework
on semantically labeled point cloud maps
and semantic segmentation images as queries.
Leveraging the efficient graph-theoretic
correspondence matching with
the simple consistency criteria,
we presented promising preliminary results
on the global localization task
in simulated urban environments.
This research direction will lead to
localization ability of robots with
more flexibility of 
of the map and sensor modalities.




%




\printbibliography

\end{document}